\documentclass[conference]{IEEEtran}
\IEEEoverridecommandlockouts
\usepackage{cite}
\usepackage{amsmath,amssymb,amsfonts}
\usepackage{algorithmic}
\usepackage{graphicx}
\usepackage{textcomp}
\usepackage{xcolor}
\usepackage{todonotes}
\def\BibTeX{{\rm B\kern-.05em{\sc i\kern-.025em b}\kern-.08em
    T\kern-.1667em\lower.7ex\hbox{E}\kern-.125emX}}
\begin{document}

\newcommand\copyrighttext{%
  \footnotesize \textcopyright 2021 IEEE. Personal use of this material is permitted. Permission from IEEE must be obtained for all other uses, in any current or future media, including reprinting/republishing this material for advertising or promotional purposes,creating new collective works, for resale or redistribution to servers or lists, or reuse of any copyrighted component of this work in other works.\\
  Conference: 2021 IEEE Ural Symposium on Biomedical Engineering, Radioelectronics and Information Technology (USBEREIT 2021)}
\newcommand\copyrightnotice{%
\begin{tikzpicture}[remember picture,overlay]
\node[anchor=south,yshift=10pt] at (current page.south) {\fbox{\parbox{\dimexpr\textwidth-\fboxsep-\fboxrule\relax}{\copyrighttext}}};
\end{tikzpicture}%
}

\newpage

\title{Anomaly Detection in Image Datasets Using Convolutional Neural Networks,  Center Loss, and Mahalanobis Distance}

\author{\IEEEauthorblockN{Garnik Vareldzhan}
\IEEEauthorblockA{\textit{R\&D Department} \\
\textit{OOO Code Laboratory}\\
Perm, Russian Federation \\
gavid19912@gmail.com}
\and
\IEEEauthorblockN{Kirill Yurkov}
\IEEEauthorblockA{\textit{Mechanics and Mathematics} \\
\textit{Perm State National Research University }\\
Perm, Russian Federation \\
opcheese@gmail.com}
\and
\IEEEauthorblockN{Konstantin Ushenin}
\IEEEauthorblockA{\textit{Mathematical Physiology} \\
\textit{Institute of Immunology and Physiology}\\
Ekaterinburg, Russian Federation \\
konstantin.ushenin@urfu.ru}
}

\maketitle
\copyrightnotice

\begin{abstract}
User activities generate a significant number of poor-quality or irrelevant images and data vectors that cannot be processed in the main data processing pipeline or included in the training dataset. Such samples can be found with manual analysis by an expert or with anomalous detection algorithms. There are several formal definitions for the anomaly samples. For neural networks, the anomalous is usually defined as out-of-distribution samples. This work proposes methods for supervised and semi-supervised detection of out-of-distribution samples in image datasets. Our approach extends a typical neural network that solves the image classification problem. Thus, one neural network after extension can solve image classification and anomalous detection problems simultaneously. Proposed methods are based on the center loss and its effect on a deep feature distribution in a last hidden layer of the neural network. This paper provides an analysis of the proposed methods for the LeNet and EfficientNet-B0 on the MNIST and ImageNet-30 datasets.
\end{abstract}



\section{Introduction}

Convolution neural networks as an image processing approach show the best performance for the image classification problem. However, the usage of neural networks in real-life applications has a number of practical challenges. Primarily, these challenges are related to datasets: small size of the training dataset, train-test leakage,  sample imbalance, and others. Other problems are related to user behavior. For example, these are irrelevant and poor quality of user-produced data.

Detection of unusual, irrelevant, or adversarial data is important both for the creation of a proper training dataset and for filtering irrelevant samples during the inference. Several formal definitions are proposed for these problems, such as anomaly detection, novelty detection, outlier detection, or out-of-distribution detection. In this work, we use terms definitions from \cite{bulusu2020anomalous}. An anomalous detection is a general term for the detection of any unusual or unwanted data. An out-of-distribution (OOD) detection is a more specific problem. This is the detection of samples that are included in the distribution of test samples (inference samples) but are not included in the distribution of training samples. OOD detection problem assumes that detected samples unintentionally appear in the test dataset and their inclusion in the training dataset is not required after detection. That problem focus differentiates OOD detection from adversarial attack detection and novelty detection, respectively.

According to \cite{bulusu2020anomalous}, OOD detection approaches may be trained with supervised, semi-supervised, or unsupervised learning. In the first case, the algorithm observes in-distribution and OOD samples with proper markup. Thus, ODD detection is equivalent to a binary classification problem. Semi-supervised learning uses only the in-distribution samples. Unsupervised learning does not use any data markup. Evaluation of OOD detection methods requires the main dataset with in-distribution samples and anomalous dataset with proper out-of-distribution samples.

There are a lot of methods for anomaly and OOD detection in vector datasets: the local outlier factor, Mahalanobis distance, isolation forest, one-class support vector machine, autoencoder, variational autoencoder \cite{masaki2020anomaly,pol2019anomaly}, gaussian mixture model for variational autoencoder, and others. Most of them are implemented in open-source libraries \cite{scikit-learn, alibi-detect}. Methods for anomaly detection in image datasets are also proposed. For example, this is a convolutional autoencoder. Recent studies propose methods based on contrastive learning \cite{tack2020csi}, and variations of a RotNet \cite{hendrycks2019using}.

All listed approaches focus only on anomalous or ODD detection problems. However, these problems are rarely the main goals of data processing and, usually, they are only small parts of a bigger data processing workflow. Neural networks are a computationally expensive approach. For this reason, an extension of neural networks is preferable to adding one more neural network to the workflow. 

This study proposes two methods which extend convolutional neural networks for image classification problem. The extended networks solve classification and OOD detection problems simultaneously. Thus, methods reduce time and computational cost for the training and inference of neural networks. The first proposed method is a semi-supervised approach that combines Mahalanobis distance and a center loss \cite{wen2016discriminative}. The second method is a supervised OOD detection approach based on multi-layer perception as the second head of the neural network.  Both approaches were evaluated with MNIST \cite{deng2012mnist}, FashionMNIST \cite{xiao2017fashion}, and ImageNet30 \cite{hendrycks2019using} datasets using LeNet \cite{lecun2015lenet} and EfficientNet-B0 \cite{tan2019efficientnet} neural networks.

\section{Methods}

\begin{figure}[!h]
\centering
\includegraphics[width=0.5\textwidth]{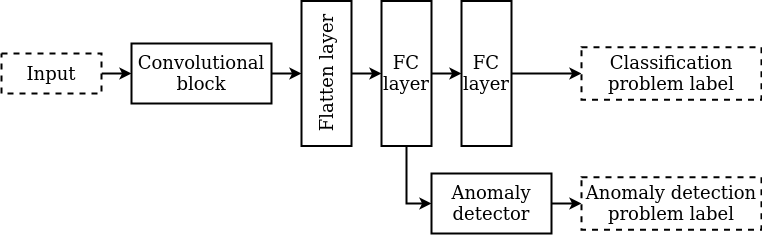}
\caption{The typical convolutional neural network for image classification.}\label{fig:main_scheme}
\end{figure}

Fig. \ref{fig:main_scheme} shows a typical convolutional network that solves a classification problem \cite{zhang2020dive}. The neural network obtains a mini-batch of images $\{\mathbf{X}_i\}_{i=0}^m$ and produces a set of predicted classes $\{\hat{y}_i\}_{i=0}^m$. A set of true classes is $\{y_i\}_{i=0}^m$. The last hidden layer provides a set of deep feature vectors $\{\mathbf{x}_i\}_{i=0}^m$. Our approaches to OOD detection extends such neural networks.

\textit{Semi-supervised OOD detection method} compliments cross-entropy loss with a center loss:
\begin{align}
    \mathcal{L} &= \mathcal{L}_S + \lambda \mathcal{L}_C = \\ &= - \sum^{m}_{i=1} \log \frac{e^{W^T_{y_i}\mathbf{x}_i+b_{y_i}}}{\sum^{n}_{j=1} e^{W^T_j\mathbf{x}_i+b_{j}}} + \frac{\lambda}{2}\sum^{m}_{i=1}\|\mathbf{x}_i - \mathbf{c}_{y_i}\|^2_2,
\end{align}
where $L$ denotes a complex loss function; $\mathcal{L}_S$ denotes a softmax function with a cross entropy; $\mathcal{L}_C$ denotes the center loss function; $\lambda$ is a balancing coefficient; $n$ and $m$ denote a number of class and a size of mini-batch, respectively; $\mathbf{x}_i \in \mathbb{R}^d$ denotes the $i$th deep feature, beloning to the $y_i$th class; $d$ is a feature dimentions; $W_j \in \mathbb{R}^d$ denotes the $j$th column of the weights $W \in \mathbb{R}^{d \times n}$ in the last fully connected layer and $\mathbf{b} \in \mathbb{R}^n$ is the bias term; $\mathbf{c}_{y_i} \in \mathbb{R}^d$ denotes the $y_i$th class center of deep features. 

Training of neural network for classification problem is performed with the main training dataset. After the training, all samples $\mathbf{X}_i,\ i \in [1, M],\ i\in\mathbb{N}$ passed to the neural network and this generates deep feature vectors $\mathbf{x}_i = T(\mathbf{X}_i)$. Each vector belongs to one given class $\{(\mathbf{x}_i,y_i)\}_{i=1}^{M}$. Let us denote subset of deep features that relates to $y_i$th class as $\{\mathbf{x}\}^{(y_i)}$. Each subset of deep features determines a mean vector $\boldsymbol{\mu}^{(y_i)} = \operatorname {E}[\{\mathbf{x}\}^{(y_i)}]$ and covariance matrix $\mathbf{S}^{(y_i)}$. Obtained vectors determine Mahalanobis distances for each class: $D_{M}^{(y_i)}(\mathbf{x})=\sqrt{(\mathbf{x}-\boldsymbol{\mu}^{(y_i)})^T (\mathbf{S}^{(y_i)})^{-1} (\mathbf{x}-\boldsymbol{\mu}^{(y_i)})}$. Distances of deep features obtained for $y_i$th determines $\boldsymbol{\theta}^{(y_i)}$ that is a threshold criteria. We choose threshold as 97.5\% percentile: $\boldsymbol{\theta}^{(y_i)} = \operatorname{percentile}(\{ D_{M}^{(y_i)}(\mathbf{x}) | \mathbf{x} \in \{\mathbf{x}\}^{(y_i)} \}, 0.975)$.

Training with center loss provides a deep feature space, where $\mathbf{x}^{(y_i)}$ groups around proper centroids. We assume that OOD samples $\mathbf{X}^{(-1)}$ transformed to deep feature vector $\mathbf{x}^{(-1)}$ that are far from any centroids. The following criteria are used to differentiate OOD and normal sample:
\begin{align}
    P(\mathbf{X}) &= Q(T(\mathbf{X})),  \\
    Q(\mathbf{x}) &= \bigvee_{i=1}^n (D_{M}^{(y_i)}(\mathbf{x}) \leq \boldsymbol{\theta}^{(y_i)}),\label{eq:criteria} \\
    P,Q &\in \{\operatorname{False}, \operatorname{True}\},
\end{align}
where  $P(\mathbf{X}) \equiv \operatorname{False}$ for OOD samples, and $P(\mathbf{X}) \equiv \operatorname{True}$ for normal samples.

\textit{Supervised OOD detection method} also uses the center loss but analysis of the deep features vectors are performed with the multi-layer perceptron instead of the Mahalanobis distance criteria. The multi-layer perceptron includes three fully connected layers with 256, 256, and 1 neuron respectively. Two layers use ReLU as the activation function, and the last layer uses the sigmoid function. The second head is trained with the binary cross-entropy loss function.

Training of the neural network is performed in two-stage. The first stage is training for the classification problem with the main training dataset. The backpropagation algorithm here uses complex loss ($\mathcal{L} = \mathcal{L}_S + \lambda \mathcal{L}_C$) and propagates a prediction error from the main head to the general part of the neural network. The second stage is a training of the second head for OOD prediction, using samples from the anomalous dataset as the zeros class and samples from the main dataset as the first class. This problem is equivalent to binary classification, where the zeroth class is OOD samples, and the first class is normal samples.

\textit{Analysis of the proposed method} were performed using modified LeNet \cite{lecun2015lenet} and EfficientNet-B0 \cite{tan2019efficientnet} neural networks. The modified LeNet have the same structure as the network from the original work \cite{lecun2015lenet}, but we replace average pooling layers with max-pooling layers and replace hyperbolic tangent activation functions with rectifier linear unit functions \cite{zhang2020dive}. These changes aim to increase of the training and inference performance. The EfficientNet-B0 \cite{tan2019efficientnet} were adopted from \cite{tan2019efficientnet} without change. 

This study uses datasets of the handwritten digest (MNIST), images of clothing (FashionMNIST), and images of real-world objects (ImageNet30, \cite{hendrycks2019using}). The MNIST and FashionMNIST consist of grayscale images with the size of 28x28 px. The ImageNet30 dataset consists of RGB-images with size of 256x256 px. Datasets of main data and anomaly data were obtained using splitting of MNIST dataset into MNIST-0 and MNIST-1..9 datasets according to the zero and other digits. MNIST and MNIST-1..9 were used as the main datasets, the FashionMNIST and MNIST-0 as the anomaly dataset. We experimented with 3 different splits of ImageNet-30: ImageNet-30a consists of images from 0 to 9 classes, ImageNet-30b consists of images from 10 to 19, and  ImageNet-30c consists of images from 10 to 19. In computational experiments, one of three datasets become the main dataset and another one becomes the anomalous dataset.
\section{Results}
\begin{figure}[h]
\centering
\includegraphics[width=0.45\textwidth]{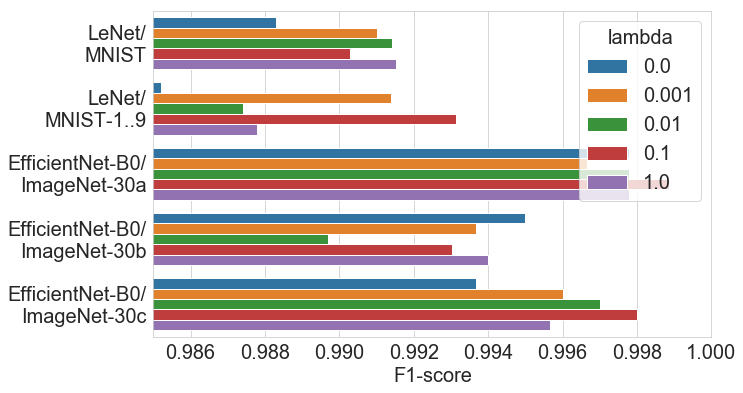}
\caption{Performance of neural networks in the classification problem for some values of the balancing coefficient $\lambda$.}  \label{fig:f1-classification}
\end{figure}

\begin{figure}[h]
\centering
\includegraphics[width=0.45\textwidth]{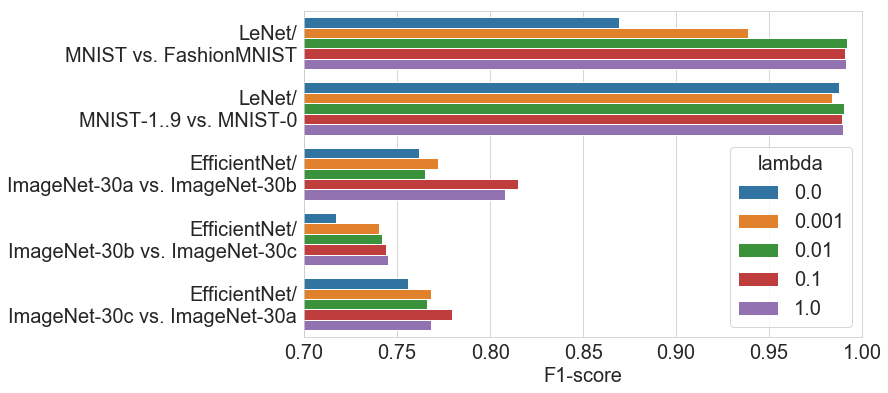}
\caption{Performance of neural networks in OOD sample detection using semi-supervised approach for some values of the balancing coefficient $\lambda$.}  \label{fig:f1-Mahalanobis}
\end{figure}

\begin{figure}[h]
\centering
\includegraphics[width=0.45\textwidth]{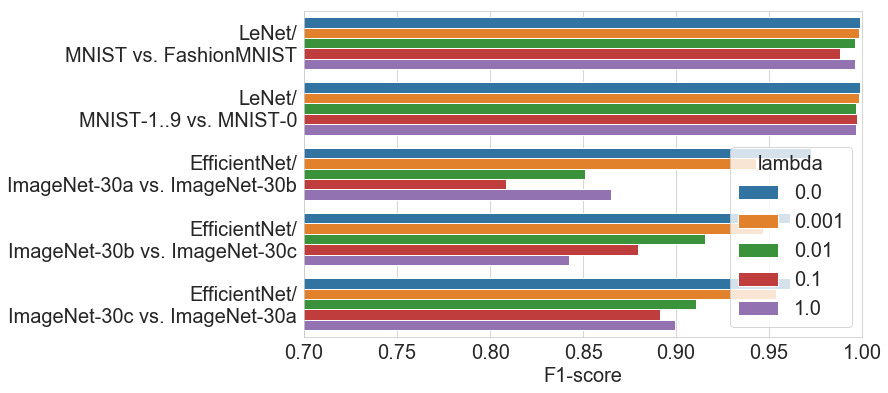}
\caption{Performance of neural networks in ODD sample detection using supervised approach for some values of the balancing coefficient $\lambda$.}  \label{fig:f1-MLP}
\end{figure}

Fig. \ref{fig:f1-classification} shows F1-score for the classification problem. The LeNet network on MNIST and MNIST-1..9 showed the F1-score in a range from 0.9852 to 0.9915. The EfficientNet-B0 on the ImageNet-30a, ImageNet-30b, ImageNet-30c showed F1-score in range from 0.9897 to 0.9989. The performance of EfficientNet in the last case was better because this neural network includes more hidden layers. The value of the balancing coefficient $\lambda$ significantly affected the classification accuracy and F1-score. Usage of the center loss improved the classification accuracy of the LeNet in all experiments. The center loss also improved results for ImageNet-30 in one of three experiments. The worst decrease of classification accuracy and F1-score caused by center loss did not overcome 0.0052 and 0.005295 respectively. Thus, we suppose that usage of the center loss is appropriate in real-life applications. According to ranges of F1-scores, the optimal value of the balancing coefficient $\lambda$ was 0.1 or 1.
Fig. \ref{fig:f1-Mahalanobis} shows the performance of anomaly detection with the proposed semi-supervised OOD detection method. As can be seen, the center loss improved the F1-score in all five experiments. The F1-score raised more than 0.03 points in three cases.  According to ranges of F1-scores, the optimal value of the balancing coefficient $\lambda$ is 0.1 or 1.
Analysis of results for supervised classification provided opposite results. As shown in Fig. \ref{fig:f1-MLP}, the best F1-score for OOD detection was reached without the center loss ($\lambda=0.0$). The OOD detector performance was decreased by 0.19 points in the worst case. This effect is more significant for datasets with real-world objects and for deeper neural networks. 

\begin{figure}[h]
\centering
\includegraphics[width=0.45\textwidth]{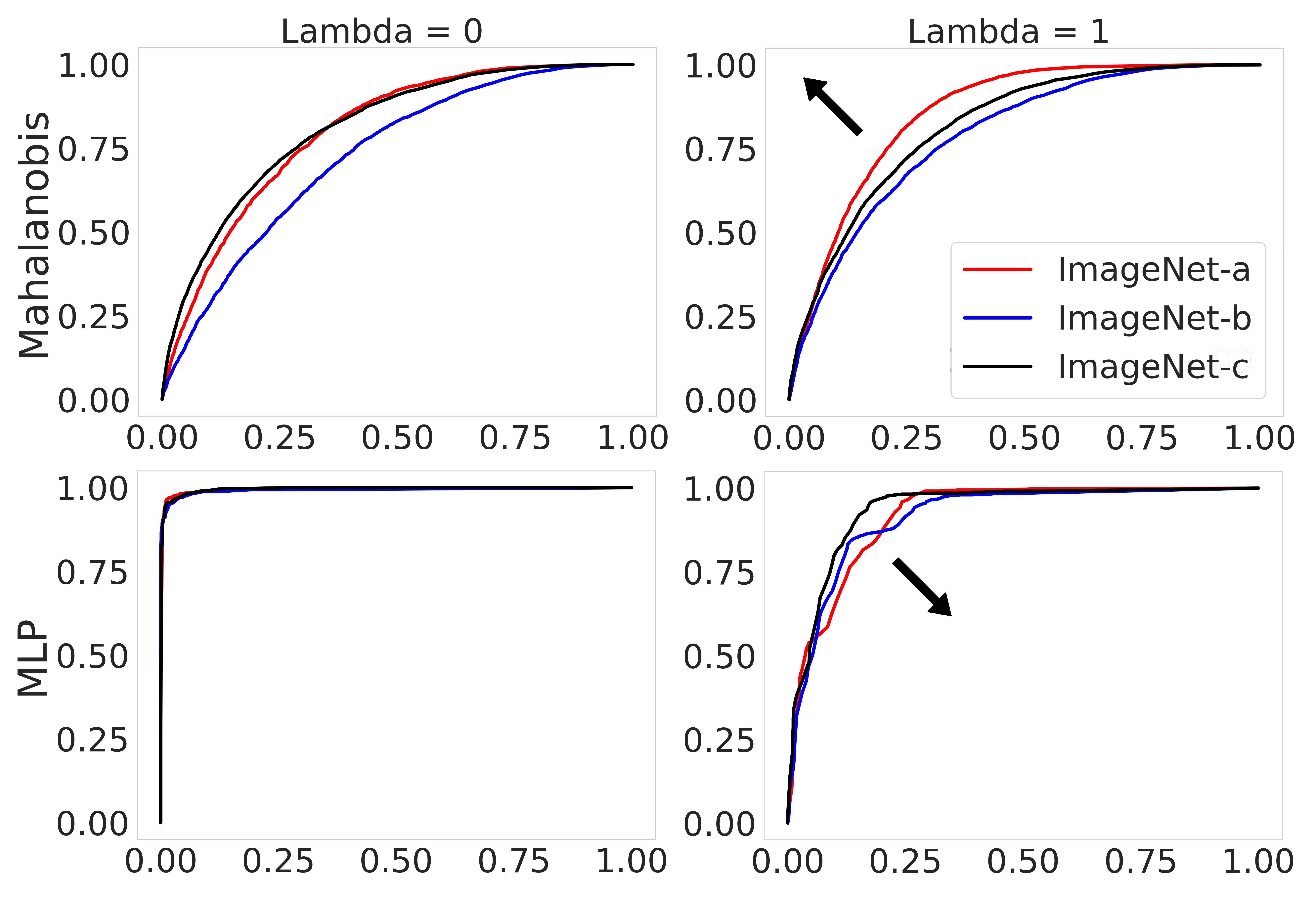}
\caption{Effect of the balancing coefficient $\lambda$ on the ROC curves for semi-supervised approach (Mahalanobis) and supervised approach (MLP). Black arrows show direction of changes that observed for $\lambda$ increasing.}  \label{fig:ROC-curve}
\end{figure}

OOD detection for the supervised method was better than for the unsupervised. F1-score for semi-supervised approach was in range $[0.8695, 0.9913]$ for the LeNet and was range $[0.7172, 0.8149]$ for the EfficientNet. Supervised approach shows F1-score in ranges $[0.9881, 0.9990]$ and $[0.8083, 0.9727]$, respectively. Thus, we conclude that the supervised method is preferable in real applications.
According to obtained results, the center loss is suitable for the semi-supervised method, but cannot bring advantages in supervised OOD detection. These were be studied more precisely using ROC curve analysis. Fig. \ref{fig:ROC-curve} shows ROC curve for both methods under $\lambda=0$ or $\lambda=1$. ROC curves for semi-supervised method were built by simultaneous variation of all thresholds $\boldsymbol{\theta}^{(y_i)}$ in range $[0,6]$. ROC curves for the supervised method were obtained using a variety of thresholds in the sigmoid function. The plots show that the increase of $\lambda$ affected two methods in opposite directions. High $\lambda$ improved the semi-supervised approach, but the center loss decreased area under the curve for the supervised one.
\begin{figure*}[!t]
\centering
\includegraphics[width=\textwidth]{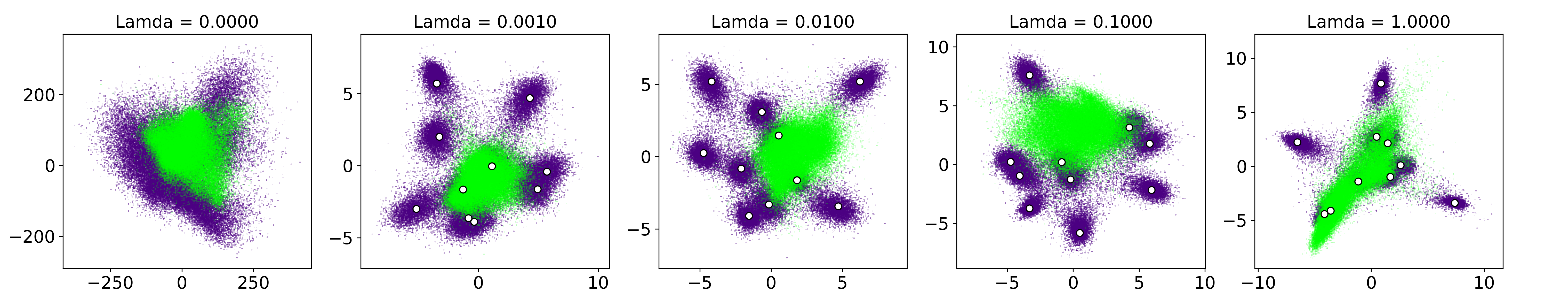}
\caption{Multidimensional deep feature vectors that are represented on 2D plane. This projection is obtained using the principal component analysis. Green dots are deep feature vectors transformed from OOD samples. Blue dots are deep feature vectors transformed from normal samples. White circles are centroid locations.}  \label{fig:lambda_picture_1}
\end{figure*}
Above in the text, we introduce criteria for OOD detection assuming that normal samples are localized near the centroids. Fig. \ref{fig:lambda_picture_1} proves this assumption. Without centroids OOD and normal samples form an unstructured distribution with mean near to the zero vector (see Fig. \ref{fig:lambda_picture_1} for $\lambda=0$). Usage of the center loss modified distributions of the deep feature vectors. OOD samples transformed to deep vectors $\mathbf{x}^{(-1)}$ that surrounding the zero vector, but normal samples are distributed around centroids.
\section{Discussion}

In this study, we propose supervised and semi-supervised methods for anomaly detection in image dataset (out-of-distribution samples detection). The proposed approaches extend a typical convolutional neural network that solves the problem of image classification. Thus, the modified network performs classification and OOD detection simultaneously. The semi-supervised approach is based on the usage of the center loss to build the suitable distribution of deep vectors and usage of Mahalanobis distance for analysis of this distribution. The supervised approach is based on the second head that analyzing deep features. 

The semi-supervised method is more agile because it does not require a dataset with anomaly samples for training. Usage of the center loss affects the image classification accuracy insignificantly but strongly affects the deep feature distributions. Normal samples are grouped near the centroids while out-of-distribution samples are grouped near the zero vector. Then, criteria (\ref{eq:criteria}) separates OOD and normal samples from image datasets. Increasing the balancing coefficient $\lambda$ improves the accuracy and area under the ROC curve for OOD detection.

The supervised approach shows better performance than the semi-supervised. That makes it favorable for real application. However, this approach is strongly dependent on the choice of a proper anomalous dataset. The center loss should not be used with the supervised approach. 

\bibliographystyle{IEEEtran}
\bibliography{main}

\begin{thebibliography}{10}
\providecommand{\url}[1]{#1}
\csname url@samestyle\endcsname
\providecommand{\newblock}{\relax}
\providecommand{\bibinfo}[2]{#2}
\providecommand{\BIBentrySTDinterwordspacing}{\spaceskip=0pt\relax}
\providecommand{\BIBentryALTinterwordstretchfactor}{4}
\providecommand{\BIBentryALTinterwordspacing}{\spaceskip=\fontdimen2\font plus
\BIBentryALTinterwordstretchfactor\fontdimen3\font minus
  \fontdimen4\font\relax}
\providecommand{\BIBforeignlanguage}[2]{{%
\expandafter\ifx\csname l@#1\endcsname\relax
\typeout{** WARNING: IEEEtran.bst: No hyphenation pattern has been}%
\typeout{** loaded for the language `#1'. Using the pattern for}%
\typeout{** the default language instead.}%
\else
\language=\csname l@#1\endcsname
\fi
#2}}
\providecommand{\BIBdecl}{\relax}
\BIBdecl

\bibitem{bulusu2020anomalous}
S.~Bulusu, B.~Kailkhura, B.~Li, P.~K. Varshney, and D.~Song, ``Anomalous
  instance detection in deep learning: A survey,'' \emph{arXiv preprint
  arXiv:2003.06979}, 2020.

\bibitem{masaki2020anomaly}
A.~Masaki, K.~Nagumo, B.~Lamsal, K.~Oiwa, and A.~Nozawa, ``Anomaly detection in
  facial skin temperature using variational autoencoder,'' \emph{Artificial
  Life and Robotics}, pp. 1--7, 2020.

\bibitem{pol2019anomaly}
A.~A. Pol, V.~Berger, C.~Germain, G.~Cerminara, and M.~Pierini, ``Anomaly
  detection with conditional variational autoencoders,'' in \emph{2019 18th
  IEEE International Conference On Machine Learning And Applications
  (ICMLA)}.\hskip 1em plus 0.5em minus 0.4em\relax IEEE, 2019, pp. 1651--1657.

\bibitem{scikit-learn}
F.~Pedregosa, G.~Varoquaux, A.~Gramfort, V.~Michel, B.~Thirion, O.~Grisel,
  M.~Blondel, P.~Prettenhofer, R.~Weiss, V.~Dubourg, J.~Vanderplas, A.~Passos,
  D.~Cournapeau, M.~Brucher, M.~Perrot, and E.~Duchesnay, ``Scikit-learn:
  Machine learning in {P}ython,'' \emph{Journal of Machine Learning Research},
  vol.~12, pp. 2825--2830, 2011.

\bibitem{alibi-detect}
A.~Van~Looveren, G.~Vacanti, J.~Klaise, and A.~Coca, ``{Alibi-Detect}:
  Algorithms for outlier and adversarial instance detection, concept drift and
  metrics.'' 2019.

\bibitem{tack2020csi}
J.~Tack, S.~Mo, J.~Jeong, and J.~Shin, ``Csi: Novelty detection via contrastive
  learning on distributionally shifted instances,'' \emph{arXiv preprint
  arXiv:2007.08176}, 2020.

\bibitem{hendrycks2019using}
D.~Hendrycks, M.~Mazeika, S.~Kadavath, and D.~Song, ``Using self-supervised
  learning can improve model robustness and uncertainty,'' in \emph{Advances in
  Neural Information Processing Systems}, 2019, pp. 15\,663--15\,674.

\bibitem{wen2016discriminative}
Y.~Wen, K.~Zhang, Z.~Li, and Y.~Qiao, ``A discriminative feature learning
  approach for deep face recognition,'' in \emph{European conference on
  computer vision}.\hskip 1em plus 0.5em minus 0.4em\relax Springer, 2016, pp.
  499--515.

\bibitem{deng2012mnist}
L.~Deng, ``The mnist database of handwritten digit images for machine learning
  research [best of the web],'' \emph{IEEE Signal Processing Magazine},
  vol.~29, no.~6, pp. 141--142, 2012.

\bibitem{xiao2017fashion}
H.~Xiao, K.~Rasul, and R.~Vollgraf, ``Fashion-mnist: a novel image dataset for
  benchmarking machine learning algorithms,'' \emph{arXiv preprint
  arXiv:1708.07747}, 2017.

\bibitem{lecun2015lenet}
Y.~LeCun \emph{et~al.}, ``Lenet-5, convolutional neural networks,'' \emph{URL:
  http://yann. lecun. com/exdb/lenet}, vol.~20, no.~5, p.~14, 2015.

\bibitem{tan2019efficientnet}
M.~Tan and Q.~V. Le, ``Efficientnet: Rethinking model scaling for convolutional
  neural networks,'' \emph{arXiv preprint arXiv:1905.11946}, 2019.

\bibitem{zhang2020dive}
A.~Zhang, Z.~C. Lipton, M.~Li, and A.~J. Smola, \emph{Dive into Deep Learning},
  2020, \url{https://d2l.ai}.

\end{thebibliography}
\end{document}